\documentclass{article}

\usepackage[final]{nips_2017}

\usepackage[utf8]{inputenc} % allow utf-8 input
\usepackage[T1]{fontenc}    % use 8-bit T1 fonts
\usepackage{hyperref}       % hyperlinks
\usepackage{url}            % simple URL typesetting
\usepackage{booktabs}       % professional-quality tables
\usepackage{amsfonts}       % blackboard math symbols
\usepackage{nicefrac}       % compact symbols for 1/2, etc.
\usepackage{microtype}      % microtypography
\usepackage{graphicx}
\usepackage{amsmath}
\usepackage{url}
\usepackage{multirow}
\usepackage{hhline}

\title{Learning Word Embeddings from Speech}

\author{
  Yu-An~Chung and James Glass\\
  Computer Science and Artificial Intelligence Laboratory\\
  Massachusetts Institute of Technology\\
  Cambridge, MA 02139\\
  \texttt{\{andyyuan,glass\}@mit.edu}\\
}

\begin{document}

\maketitle

\begin{abstract}
  In this paper, we propose a novel deep neural network architecture, Sequence-to-Sequence Audio2Vec, for unsupervised learning of fixed-length vector representations of audio segments excised from a speech corpus, where the vectors contain semantic information pertaining to the segments, and are close to other vectors in the embedding space if their corresponding segments are semantically similar.
  The design of the proposed model is based on the RNN Encoder-Decoder framework, and borrows the methodology of continuous skip-grams for training.
  The learned vector representations are evaluated on~13 widely used word similarity benchmarks, and achieved competitive results to that of GloVe.
  The biggest advantage of the proposed model is its capability of extracting semantic information of audio segments taken directly from raw speech, without relying on any other modalities such as text or images, which are challenging and expensive to collect and annotate.
\end{abstract}

\section{Introduction}
Natural language processing~(NLP) techniques such as GloVe~\citep{pennington2014glove} and word2vec~\citep{mikolov2013distributed} transform words into fixed dimensional vectors.
The vectors are obtained by unsupervised learning from co-occurrences information in the text, and contain semantic information about the word which are useful for many NLP tasks.
Given the observation that humans learn to speak before they can read or write, one might wonder that since machines can learn semantics from raw text, might they also be able to learn the semantics of a spoken language from raw speech as well? 

Previous research has explored the concept of learning vector representations from speech~\citep{he2017multi,kamper2016deep,chung2016audio,settle2016discriminative,bengio2014word,levin2013fixed}.
These approaches were based on notions of acoustic-phonetic similarity, rather than {\it semantic}, so that different instances of the same underlying word would map to the same point in the embedding space.
Our work uses a very different skip-gram formulation to focus on the semantics of {\it neighboring} acoustic regions, rather than acoustic segment associated with the word itself.

Recent research by~\cite{harwath2017learning,harwath2016unsupervised,harwath2015deep} has presented a deep neural network model capable of rudimentary spoken language acquisition using raw speech training data paired with contextually relevant images.
Using this contextual grounding, the model learned a latent semantic audio-visual embedding space.
In this paper, we propose a deep neural network architecture capable of learning fixed-length vector representations of audio segments from \textit{raw} speech without any other modalities, such that the vector representations contain semantic information of underlying words.
The proposed model, called Sequence-to-Sequence Audio2Vec, integrates an RNN Encoder-Decoder framework with the concept of continuous skip-grams, and can handle arbitrary length speech segments.
The resulting vector representations contain information pertaining to the meaning of the underlying spoken words such that semantically similar words produce vector representations that are nearby in the embedding space.

\section{Proposed Approach}
Our goal is to learn a fixed-length vector representation of an audio segment that is represented by a variable-length sequence of acoustic features such as Mel-Frequency Cepstral Coefficients~(MFCCs), $\mathbf{x} = (\mathbf{x}_{1}, \mathbf{x}_{2}, ..., \mathbf{x}_{T})$, where~$\mathbf{x}_{t}$ is the acoustic feature at time~$t$ and~$T$ is the length of the sequence.
We desire that this fixed-length vector representation is able to describe the semantics of the original audio segment to some degree.
Below we first review the RNN Encoder-Decoder framework in Section~\ref{sec:encoder-decoder}, followed by formally proposing the Sequence-to-Sequence Audio2Vec model in Section~\ref{sec:audio2vec}.

\subsection{RNN Encoder-Decoder Framework}
\label{sec:encoder-decoder}
Recurrent neural networks~(RNNs) are neural networks whose hidden neurons form a directed cycle.
Given a sequence~$\mathbf{x} = (\mathbf{x}_{1}, \mathbf{x}_{2}, ..., \mathbf{x}_{T})$, an RNN updates its hidden state~$\mathbf{h}_{t}$ according to the current input~$\mathbf{x}_{t}$ and the previous~$\mathbf{h}_{t - 1}$.
The hidden state~$\mathbf{h}_{t}$ acts as a form of internal memory at time~$t$ that enables the network to capture dynamic temporal information, and also allows the network to process variable length sequences.
Unfortunately, in practice RNNs do not seem to learn long-term dependencies well, so Long Short-Term Memory~(LSTM) networks~\citep{hochreiter1997long}, an advanced version of the vanilla RNN, have been widely used to conquer such difficulties.

An RNN Encoder-Decoder consists of an Encoder RNN and a Decoder RNN~\citep{sutskever2014sequence,cho2014learning}.
The Encoder reads the input sequence~$\mathbf{x} = (\mathbf{x}_{1}, \mathbf{x}_{2}, ..., \mathbf{x}_{T})$ sequentially, and the hidden state~$\mathbf{h}_{t}$ of the RNN is updated accordingly.
After the last symbol~$\mathbf{x}_{T}$ is processed, the corresponding hidden state~$\mathbf{h}_{T}$ is interpreted as the learned representation of the entire input sequence.
Subsequently, by initializing its hidden state using~$\mathbf{h}_{T}$, the Decoder generates an output sequence~$\mathbf{y} = (\mathbf{y}_{1}, \mathbf{y}_{2}, ..., \mathbf{y}_{T'})$ sequentially, where~$T$ and~$T'$ can be different, or, in other words, the sequence lengths of~$\mathbf{x}$ and~$\mathbf{y}$ can be different.
Such a sequence-to-sequence framework does not constrain the input or target sequences, and has been successfully applied to a wide range of challenging tasks such as machine translation~\citep{sutskever2014sequence,cho2014learning}, video caption generation~\citep{venugopalan2015sequence}, abstract meaning representation~(AMR) parsing and generation~\citep{konstas2017neural}, and acquisition of acoustic word embeddings~\citep{chung2016audio}.

\subsection{Sequence-to-Sequence Audio2Vec}
\label{sec:audio2vec}
\begin{figure}[htbp]
  \centering
  \includegraphics[scale=0.3]{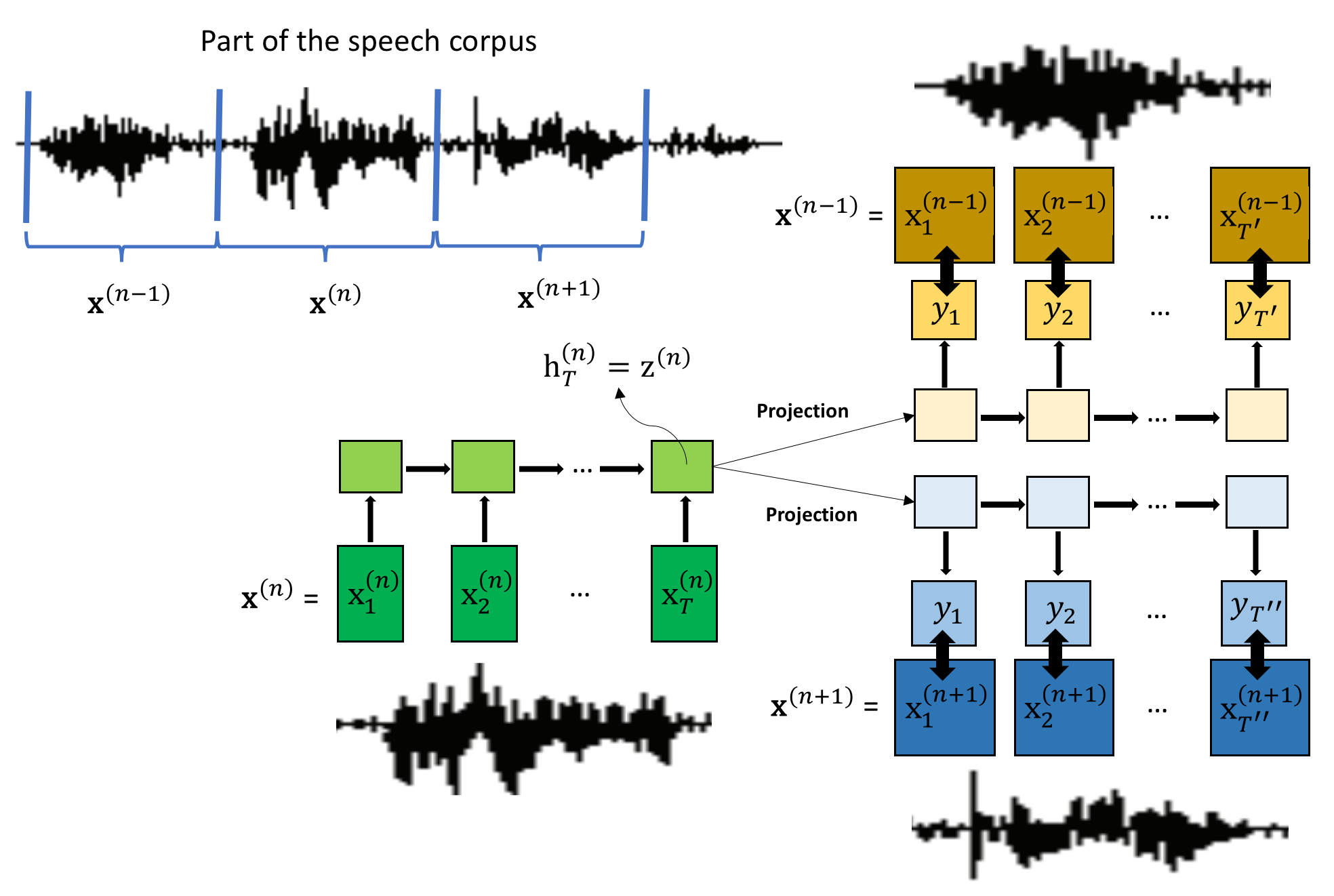}
  \caption{
  The Seq2seq Audio2vec model consists of an Encoder RNN and a Decoder RNN.
  The Encoder first takes an audio segment~$\mathbf{x}^{(n)} = (\mathbf{x}_{1}^{(n)}, \mathbf{x}_{2}^{(n)}, ..., \mathbf{x}_{T}^{(n)})$ as input and encodes it into a vector representation of fixed dimensionality~$\mathbf{z}^{(n)}$.
  The Decoder then maps~$\mathbf{z}^{(n)}$ to several audio segments~$\mathbf{x}^{(i)}, i\in \{n - k, ..., n - 1\}\bigcup \{n + 1, ..., n + k\}$ within in a certain range~$k$~(in this example,~$k = 1$).
  To successfully decode nearby audio segments, the encoded vector representation~$\mathbf{z}^{(n)}$ should contain semantic information about the current audio segment~$\mathbf{x}^{(n)}$.
  }
  \label{fig:audio2vec}
\end{figure}
Figure~\ref{fig:audio2vec} depicts the structure of the proposed Sequence-to-Sequence Audio2Vec model~(Seq2seq Audio2vec), which integrates the RNN Encoder-Decoder framework with a continuous skip-gram for unsupervised learning of audio segment representations that contain semantic information.

The idea of Seq2seq Audio2vec is simple: for each audio segment~$\mathbf{x}^{(n)}$ in a speech corpus, the model is trained to predict the audio segments~$\{\mathbf{x}^{(n - k)}, ..., \mathbf{x}^{(n - 1)}, \mathbf{x}^{(n + 1)}, ..., \mathbf{x}^{(n + k)}\}$ within a certain range~$k$ before and after~$\mathbf{x}^{(n)}$.
By applying such a methodology, the audio segments of semantically similar spoken words are mapped to nearby points in the embedding space produced by the encoder.
Figure~\ref{fig:audio2vec} is an instance of Seq2seq Audio2vec setting~$k = 1$.

The details of the proposed Seq2seq Audio2vec are as follows.
Seq2seq Audio2vec consists of an Encoder RNN and a Decoder RNN.
Given the~$n$-th audio segment in any speech corpus, represented as a sequence of acoustic features~$\mathbf{x}^{(n)} = (\mathbf{x}_{1}^{(n)}, \mathbf{x}_{2}^{(n)}, ..., \mathbf{x}_{T}^{(n)})$ of any length~$T$, the Encoder RNN reads each acoustic feature~$\mathbf{x}_{t}^{(n)}$ sequentially and updates the hidden state~$\mathbf{h}_{t}^{(n)}$ accordingly.
After the last acoustic feature~$\mathbf{x}_{T}^{(n)}$ has been read and processed, the hidden state~$\mathbf{h}_{T}^{(n)}$ of the Encoder RNN is viewed as the \textit{learned representation}~$\mathbf{z}^{(n)}$ of the current audio segment~$\mathbf{x}^{(n)}$.
The Decoder RNN now takes over the process.
It first initializes its hidden state with~$\mathbf{h}_{T}^{(n)}$, then for each audio segment~$\mathbf{x}^{(i)}, i\in \{n - k, ..., n - 1\}\bigcup \{n + 1, ..., n + k\}$ within a certain range~$k$ before and after~$\mathbf{x}^{(n)}$, the Decoder RNN generates another sequence~$\mathbf{y}^{i} = (\mathbf{y}_{1}^{(i)}, \mathbf{y}_{2}^{(i)}, ..., \mathbf{y}_{T'}^{(i)})$.
The target of the output sequence~$\mathbf{y}^{(i)}$ is set to be the corresponding audio segment~$\mathbf{x}^{(i)}$, that is, the Decoder RNN attempts to predict {\it all} of the nearby audio segments at the same time.
Note that it is the same Decoder RNN that generates all the output audio segments, and the audio segments can have different lengths.
To successfully \textit{decode} the nearby audio segments, the learned representation~$\mathbf{z}^{(n)}$ should contain sufficiently useful information about the semantics of the current audio segment~$\mathbf{x}^{(n)}$.
The model is trained by minimizing the general mean squared error~$\sum_{i\in \{n - k, ..., n - 1\}\bigcup \{n + 1, ..., n + k\}}\begin{Vmatrix}\mathbf{x}^{(i)} - \mathbf{y}^{(i)}\end{Vmatrix}^2$.

\section{Experiments}
\subsection{Experimental Setup}
We use LibriSpeech~\citep{panayotov2015librispeech}, a large corpus of read English speech, as the data for experimentation.
The corpus contains about~500 hours of broadband speech produced by~1252 speakers.
Acoustic features consisted of 13 dimensional MFCCs produced every 10ms.
The corpus was segmented according to word boundaries obtained by forced alignment with respect to the reference transcriptions, resulting in a large set of audio segments~$\{\mathbf{x}^{(1)}, \mathbf{x}^{(2)}, ..., \mathbf{x}^{(|C|)}\}$, where~$|C|$ denotes the total number of audio segments (and words) in the corpus.

The Seq2seq Audio2vec model was implemented with PyTorch.
The Encoder consists of 3-layers of LSTMs using~300 hidden units~(so the dimensionality of the learned vector representations was 300), and the Decoder was a single-layer LSTM model with~300 hidden units.
The model was trained by stochastic gradient descent without momentum, with a fixed learning rate of~$1e-3$ and~500 epochs.
We set~$k$ to~5, meaning that during training, the model took the current audio segment~$\mathbf{x}^{(n)}$ as input and attempted to predict the audio segments of the five preceding and following word segments.

After training the model, the Decoder RNN was no longer needed and could be discarded.
Each audio segment~$\mathbf{x}^{(n)}$ in the corpus was processed by the Encoder RNN, and encoded as a vector representation~$\mathbf{z}^{(n)}$ of~300 dimensions.
The vector representations representing the audio segments of the same word were then averaged to obtain a single~300-dim vector.

\subsection{Evaluation and Results}
We evaluated the vector representations learned by the proposed Seq2seq Audio2vec model on~13 different benchmarks~\citep{faruqui2014community} that have been widely used to measure word similarity.
They are: \textbf{WS-353}~\citep{yang2006verb}, \textbf{WS-353-REL}~\citep{agirre2009study}, \textbf{WS-353-SIM}, \textbf{MC-30}~\citep{miller1991contextual}, \textbf{RG-65}~\citep{rubenstein1965contextual}, \textbf{Rare-Word}~\citep{luong2013better}, \textbf{MEN}~\citep{bruni2012distributional}, \textbf{MTurk-287}~\citep{radinsky2011word}, \textbf{MTurk-771}~\citep{halawi2012large}, \textbf{YP-130}~\citep{yang2006verb}, \textbf{SimLex-999}~\citep{hill2015simlex}, \textbf{Verb-143}~\citep{baker2014unsupervised}, and \textbf{SimVerb-3500}~\citep{gerz2016simverb}.
These~13 benchmarks contain different numbers of pairs of English words that have been assigned similarity ratings by humans, and each of them tries to evaluate the word vectors in terms of different aspects.
For example, \textbf{RG-65} and \textbf{MC-30} focus on nouns, \textbf{YC-130} and \textbf{SimVerb-3500} focus on verbs, and \textbf{Rare-Word} focuses on rare-words.
We compared the vector representations learned by Seq2seq Audio2vec with GloVe trained on Wikipedia 2014.
The similarity between a given pair of words was calculated by computing the cosine similarity between their corresponding vector representations.
We then reported the Spearman's rank correlation coefficient~$\rho$ between the rankings produced by each model against the human rankings~\citep{myers1995research}.
The results were displayed in Table~\ref{tab:word-similarity}
\begin{table}[htbp]
\centering
{\renewcommand{\arraystretch}{1.1}
\caption{The Spearman's rank correlation coefficient~$\rho$ between the rankings produced by each model against the human rankings. \#(word pairs) is the number of word pairs in the dataset, and \#(not found) is the number of word pairs whose vector representations could not be found.}
\label{tab:word-similarity}
\small
\begin{tabular}{|c|c|c|c|c|c|c|}
\hline
\multirow{2}{*}{No.} & \multirow{2}{*}{Dataset} & \multirow{2}{*}{\#(word pairs)} & \multicolumn{2}{c|}{\small{Seq2seq Audio2vec}} & \multicolumn{2}{c|}{\begin{tabular}[c]{@{}c@{}}GloVe Wikipedia 2014\end{tabular}} \\ \cline{4-7} 
                     &                          &                                 & \small{\#(not found)} & $\rho$         & \small{\#(not found)}                              & $\rho$                                      \\ \hhline{|=|=|=|=|=|=|=|}
1                    & WS-353                   & 353                             & 21                    & 0.5324         & 0                                                  & 0.6054                                      \\ \hline
2                    & WS-353-REL               & 252                             & 12                    & 0.4959         & 0                                                  & 0.5725                                      \\ \hline
3                    & WS-353-SIM               & 203                             & 7                     & 0.5842         & 0                                                  & 0.6638                                      \\ \hline
4                    & MC-30                    & 30                              & 0                     & 0.6647         & 0                                                  & 0.7026                                      \\ \hline
5                    & RG-65                    & 65                              & 0                     & 0.7274         & 0                                                  & 0.7662                                      \\ \hline
6                    & Rare-Word                & 2034                            & 783                   & 0.3158         & 252                                                & 0.4118                                      \\ \hline
7                    & MEN                      & 3000                            & 122                   & 0.6877         & 0                                                  & 0.7375                                      \\ \hline
8                    & MTurk-287                & 287                             & 13                    & 0.5647         & 0                                                  & 0.6332                                      \\ \hline
9                    & MTurk-771                & 771                             & 22                    & 0.6010         & 0                                                  & 0.6501                                      \\ \hline
10                   & YP-130                   & 130                             & 0                     & 0.5173         & 0                                                  & 0.5613                                      \\ \hline
11                   & SimLex-999               & 999                             & 0                     & 0.2985         & 0                                                  & 0.3705                                      \\ \hline
12                   & Verb-143                 & 144                             & 0                     & 0.2877         & 0                                                  & 0.3051                                      \\ \hline
13                   & SimVerb-3500             & 3500                            & 126                   & 0.2023         & 2                                                  & 0.2267                                      \\ \hline
\end{tabular}%
}
\end{table}
From Table~\ref{tab:word-similarity}, we can see that the performance of the vector representations learned by Seq2seq Audio2vec is competitive to the performance of GloVe word vectors on most of the word similarity tasks.
This demonstrates that our proposed Seq2seq Audio2vec is capable of capturing semantic information from raw speech and representing it in a fixed-length vector representation, although the scores of our model were consistently lower than that obtained by GloVe.
Aside from the differences due to the speech and text training data, we believe the reason for this difference is due to the inherent variability in speech production.
Unlike textual representations, every instance of any spoken word ever uttered is different, due to vocal tract differences across speakers, speaking styles, contextual differences, and environmental conditions, to name but a few of the major influences on a speech recording.
Clearly, one of the challenges for learning semantics directly from raw speech is to derive a more robust mechanism to address these issues.
To us, what is more impressive is that many of the test scores are close.

Using word similarity tasks as the only way to measure the quality of word vectors is not perfect and can sometimes lead to incorrect inferences~\citep{faruqui2016problems,schnabel2015eval}.
In this preliminary study, we used these word similarity benchmarks to validate the effectiveness of the proposed model for learning meaningful vector representations from speech.
In the future, we will evaluate the vector representations learned by our model on other downstream NLP tasks.
It is also true, that some supervision was incorporated into the learning by using forced alignment segmentations as the basis for audio segments.
In the future, it would be interesting to explore less supervised segmentations to learn word boundaries~\citep{kamper2017learning,kamper2017segmental,kamper2015fully}.

\section{Conclusion and Future Work}
\label{sec:conclusion}
In this paper, we proposed a Seq2seq Audio2vec model for unsupervised learning of audio segment representations.
The vector representations generated by the model were evaluated on 13 commonly used word similarity benchmarks and were compared to those produced by GloVe from text data.
To the best of our knowledge, this is the first work that attempts to learn fixed-length vector representations that contain semantic information directly, and only from raw speech.
In the future, we will evaluate the vector representations on other tasks to examine their usefulness for speech and language processing.

%\subsubsection*{Acknowledgments}
%Use unnumbered third level headings for the acknowledgments. All
%acknowledgments go at the end of the paper. Do not include
%acknowledgments in the anonymized submission, only in the final paper.

\bibliography{mybib}
\bibliographystyle{abbrvnat}

\end{document}